\PassOptionsToPackage{table}{xcolor}
\documentclass{article} 
\usepackage{iclr2025_conference,times}

\usepackage{amsmath,amsfonts,bm}

\def\eqref#1{equation~\ref{#1}}

\def\1{\bm{1}}

\DeclareMathAlphabet{\mathsfit}{\encodingdefault}{\sfdefault}{m}{sl}
\SetMathAlphabet{\mathsfit}{bold}{\encodingdefault}{\sfdefault}{bx}{n}

\usepackage{hyperref}
\usepackage{url}
\usepackage{booktabs}       
\usepackage{amsfonts}       
\usepackage{nicefrac}       
\usepackage{microtype}      
\usepackage{xcolor}         
\usepackage{graphicx}
\usepackage{booktabs}
\usepackage{multirow}
\usepackage{graphicx}
\usepackage{caption}
\usepackage{subcaption}
\usepackage{verbatim}
\usepackage{arydshln}
\usepackage{array}
\usepackage{amsmath}
\usepackage{wrapfig}
\usepackage{amssymb}
\usepackage{pifont}
\usepackage{lipsum} 

\definecolor{mygray}{gray}{0.9}
\definecolor{lighterred}{rgb}{1, 0.35, 0.35}

\title{Multi-Reward as Condition for Instruction-based Image Editing}

\author{Xin Gu$^{1,2 \ddagger}$ Ming Li$^{1}$ Libo Zhang$^{2,3}$ Fan Chen$^{1}$ Longyin Wen$^{1}$ Tiejian Luo$^{2}$ Sijie Zhu$^{1,*}$ \\
$^{1} $ByteDance Inc. $^{2}$University of Chinese Academy of Sciences \\ $^{3}$Institute of Software Chinese Academy of Sciences\\
\tt\small guxin21@mails.ucas.edu.cn
}

\iclrfinalcopy 
\begin{document}

\maketitle
\let\thefootnote\relax\footnotetext{$*$ Corresponding author, sijiezhu@bytedance.com}
\let\thefootnote\relax\footnotetext{$\ddagger$ This work was done during the first author’s internship at ByteDance}

\begin{abstract}
High-quality training triplets (instruction, original image, edited image) are essential for instruction-based image editing. Predominant training datasets (e.g., InsPix2Pix) are created using text-to-image generative models (e.g., Stable Diffusion, DALL-E) which are not trained for image editing. Accordingly, these datasets suffer from inaccurate instruction following, poor detail preserving, and generation artifacts. In this paper, we propose to address the training data quality issue with multi-perspective reward data instead of refining the ground-truth image quality. 1) we first design a quantitative metric system based on best-in-class LVLM (Large Vision Language Model), i.e., GPT-4o in our case, to evaluate the generation quality from 3 perspectives, namely, instruction following, detail preserving, and generation quality. For each perspective, we collected quantitative score in $0\sim 5$ and text descriptive feedback on the specific failure points in ground-truth edited images, resulting in a high-quality editing reward dataset, i.e., RewardEdit20K. 2) We further proposed a novel training framework to seamlessly integrate the metric output, regarded as multi-reward, into editing models to learn from the imperfect training triplets. During training, the reward scores and text descriptions are encoded as embeddings and fed into both the latent space and the U-Net of the editing models as auxiliary conditions. During inference, we set these additional conditions to the highest score with no text description for failure points, to aim at the best generation outcome. 3) We also build a challenging evaluation benchmark with real-world images/photos and diverse editing instructions, named Real-Edit. Experiments indicate that our multi-reward conditioned model outperforms its no-reward counterpart on two popular editing pipelines, i.e., InsPix2Pix and SmartEdit.
Code is released at \url{https://github.com/bytedance/Multi-Reward-Editing}.
\end{abstract}

\section{Introduction}

Text instruction-based image editing provides a natural way for general users to express their requests and customize their assets easily. Predominant state-of-the-art methods for instruction-based image editing  \citep{instructpix2pix, magicbrush, hive, smartedit} follow a data-driven pipeline to finetune pre-trained diffusion models \citep{rombach2022high,ren2025videoworld, peng2024lightweight,ye2025zerodiff} with editing data triplets, i.e., (instruction, original image, edited image). Creating a high-quality dataset of the above triplets is thus essential for successful model training.

\begin{figure}[t]
	\centering
	\includegraphics[width=\linewidth]{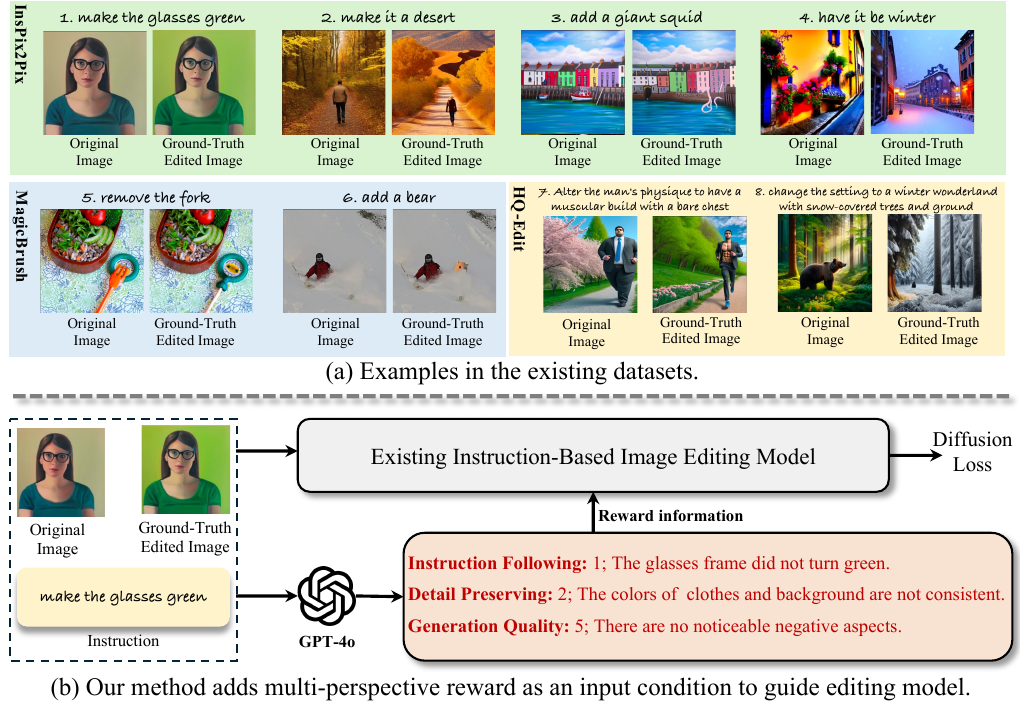}
	\caption{Existing image editing datasets and our method. Best viewed with zoom-in.}
	\label{fig:1}
    \vspace{-0.8cm}
\end{figure}

\noindent
Predominant state-of-the-art methods for instruction-based image editing  \citep{instructpix2pix, magicbrush, hive, smartedit, ho2022classifier} follow a data-driven pipeline to create the editing triplets, from which they build a dataset to fine-tune a pre-trained diffusion model \citep{rombach2022high}. The most widely used InsPix2Pix \citep{instructpix2pix} dataset is created with a pre-trained text-to-image Stable Diffusion (SD) model~\citep{rombach2022high}, Prompt-to-Prompt \citep{hertz2022prompt} and a fine-tuned GPT-3 \citep{brown2020language}. The dataset can easily scale up to 300k triplets but the quality is unsatisfactory from three perspectives, i.e., \textit{instruction following, detail preserving, and generation quality}. 
1) \textit{Instruction following} means that the model needs to closely and accurately follow the editing request, which we regard as the most important factor in instruction-based image editing. Since the SD model was originally trained for image generation tasks, it might fail to apply the correct editing action to the edited image.
As shown in Fig. \ref{fig:1} (a), the text instruction is ``make the glasses green" but the glasses in the ground-truth edited image are not green, which does not follow the major editing instruction. 
2) \textit{Detail Preserving} indicates how the model preserves identity, background or any other details that are not meant to be changed in the editing instruction. InsPix2Pix adopts prompt-to-prompt to generate edited images which could contain undesired modifications on the edited images. For example, the instruction of the first case in Fig. \ref{fig:1} (a) is to edit the color of the glasses, but the color of the clothes and background is also changed in the ground-truth edited image, which could lead to wrong supervision. 3) \textit{Generation Quality} represents the relative quality of edited images compared to the input images, i.e., to determine whether the editing action introduces quality degradation like artifacts to the real-world input images. It is common for SD models to generate artifacts, especially for images with human or small objects. In the third case of Fig. \ref{fig:1} (a), the generated ``giant squid" in the ground-truth image has serious artifacts (viewed with zoom-in). 

\noindent
MagicBrush \citep{magicbrush} leverages a more powerful text-to-image model (i.e., DALL-E 2) and human workers to improve the training data quality on a relatively small scale. The background preserving is significantly improved due to mask-based editing. However, for the edited regions inside the mask, the edited image may contain undesired modification or generation artifacts due to occlusion or small objects (see example 5,6 in Fig. \ref{fig:1}). HQEdit \citep{hui2024hq} adopts GPT-4V \citep{gpt4v} and DALL-E 3 \citep{dalle-3} to improve the instruction and generation quality. However, the edited images are usually significantly modified on the regions that are not included in the editing instruction, leading to poor detail preserving on background or identity (see example 7,8 in Fig. \ref{fig:1}). Hive \citep{hive} follows the same procedure of InsPix2Pix to create training data triplets, thus having a similar quality as InsPix2Pix. A relatively small-scale human feedback dataset is collected to improve the overall quality of the editing model, but it does not have detailed feedback information for the three perspectives of editing (i.e., following, preserving, and quality). \textit{In a nutshell, the majority of training samples in existing datasets remain noisy which could lead to inaccurate supervision.}

\noindent
In this paper, we propose to rectify the inaccurate supervision from a different perspective, i.e., introducing multi-perspective reward as an auxiliary input condition. 1) Instead of directly refining the quality of ground-truth edited images, we evaluate the training data triplets from three perspectives (i.e., instruction following, detail preserving, generation quality) with GPT-4o \citep{gpt4o} to generate scores on a 
of $0\sim 5$ and text description for unsatisfactory points. With proper prompt engineering, the generated reward/feedback is mostly aligned with humans. We collect 20k multi-perspective reward data in total for training, namely RewardEdit-20K. Examples of the scores and text description reward are included in Fig. \ref{fig:1} (b). 
2) To integrate reward information into the existing instruction-based image editing framework, we first encode the reward score and reward text description separately as embeddings, and then concatenate them to obtain the reward condition. This reward condition is then integrated into the latent noise through an attention mechanism. To further enhance the guidance provided by the reward information, we also feed the reward condition into the U-Net \citep{unet} of the SD model. 3) To evaluate the editing models on real-world photos and diverse instructions covering major 7 categories (defined in Sec. \ref{sec:eval}), we create an evaluation set with 80 high-quality Unsplash \citep{unsplash} photos and 560 challenging instructions, which are initially generated by GPT-4o and verified by human annotators. We evaluate the model output from the three perspectives with GPT-4o in terms of yes/no accuracy and score from $0\sim 5$. We also conduct a human evaluation with $0\sim 5$ score from three perspectives to further verify the results. Experiments show that the proposed method can be combined with InsPix2Pix and SmartEdit with significant performance improvement.

\noindent
We summarize the contributions as follows:
\ding{171} The RewardEdit-20K dataset with multi-perspective reward data to address the limitations of existing image editing datasets.
\ding{170} A novel framework to effectively integrate multi-perspective reward information as an additional condition to guide image editing.
\ding{169} A real-world image editing evaluation benchmark Real-Edit and introduced a GPT-4o-based image editing evaluation method.
\ding{168} Extensive experiments showing that the proposed method can be combined with existing editing models with a significant performance boost on all three perspectives, achieving state-of-the-art performance for both GPT-4o and human evaluation.

\section{Related Work}

\subsection{Instruction-based Image Editing}

Recent instruction-based image editing methods~\citep{magicbrush, peng2025towards, hive, guo2024i2v, gu2024edit3k} rely primarily on pre-trained text-to-image diffusion models. These methods leverage the powerful generative capabilities of these models and their understanding of textual descriptions to perform image editing. InsPix2Pix~\citep{instructpix2pix}, as a pioneering work, constructed a large-scale image editing dataset and successfully used instructions to edit images based on the stable diffusion model.
MagicBrush~\citep{magicbrush} addressed the issue of unrealistic images in InsPix2Pix by creating a manually annotated dataset to achieve realistic image editing.
SmartEdit~\citep{smartedit} addressed the limitation of InstructPix2Pix in handling only simple instructions by employing LLava~\citep{llava} to comprehend complex instructions.
HIVE~\citep{hive} proposed to utilize human feedback to optimize image editing models, aligning them with human preferences. \textit{However, the major training data still has a similar quality as the InsPix2Pix dataset, and the noisy supervision problem remains unaddressed.}

\subsection{Reward Mechanism for Diffusion Models}
Inspired by the success of reward fine-tuning in large language models~\citep{dpo,han2024progressivecompositionalitytexttoimagegenerative, zhang2024ctrl}, a series of works have attempted to directly optimize reward model scores~\citep{imagereward,rl_diffusion} or human preference rankings~\citep{diffusiondpo,step_dpo} to align text-to-image diffusion models, thus improving the quality, aesthetics, and text-image alignment of the generated images. 
{For text-to-image, Pony Diffusion employs a CLIP-based aesthetic ranking method to generate reward scores to improve the quality of generated images.} 
For image editing, ByteEdit~\citep{byteedit} customizes a reward model specifically for inpainting and outpainting editing tasks to identify the consistency of images beyond the mask area before and after editing. HIVE~\citep{hive}  trains a reward model to generate a single reward score for each edited image. The scores are then combined with text instructions and encoded via CLIP \citep{clip} to improve editing performance. \textit{However, there are multiple perspectives to determine the quality of an edited image given an input image and instruction, which cannot be covered by one single reward score.} Also, adding the reward scores into the text instruction does not fully exploit the reward information, as the CLIP text encoder is not sensitive to numbers. \textit{It remains challenging to effectively integrate multi-perspective reward information into existing image editing frameworks.}

\section{RewardEdit-20K: A Multi-Reward Dataset for Image Editing}

\begin{figure}[ht]
	\centering
\includegraphics[width=1.\linewidth]{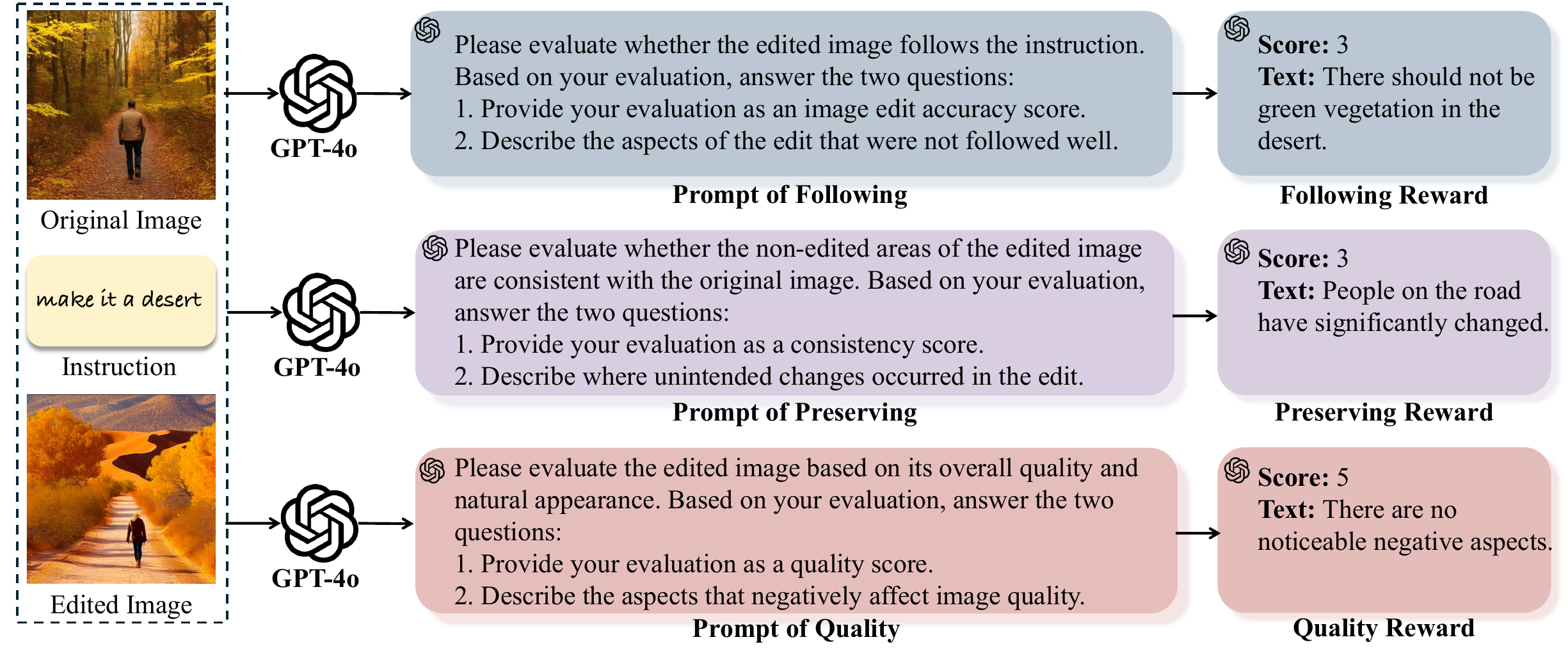}
	\caption{Generation process of reward data. Given the editing triplets, reward data was generated using GPT-4o by setting prompts from different perspectives.}
	\label{fig:generate}
\end{figure}

\textbf{Collection Process.} In this section, we discuss our procedure for collecting the RewardEdit-20K dataset. First, we randomly selected 20K training triplets from the InsPix2Pix dataset, where each triplet contains an original image, an edited image, and an editing instruction. Then, we used GPT-4o, setting up three types of prompts based on instruction following, detail preserving, and generation quality. GPT-4o was asked to perform evaluations on these three aspects for each triplet. Finally, we obtained 20K reward data consisting of reward scores and reward texts. The reward collection process is illustrated in Fig.~\ref{fig:generate}. Due to limited space, we only show the core prompts in the figure, while the complete prompts are provided in the appendix.

\begin{figure}[htb]
    \centering
    \begin{minipage}{0.33\textwidth}
        \centering
        \includegraphics[width=1.\linewidth]{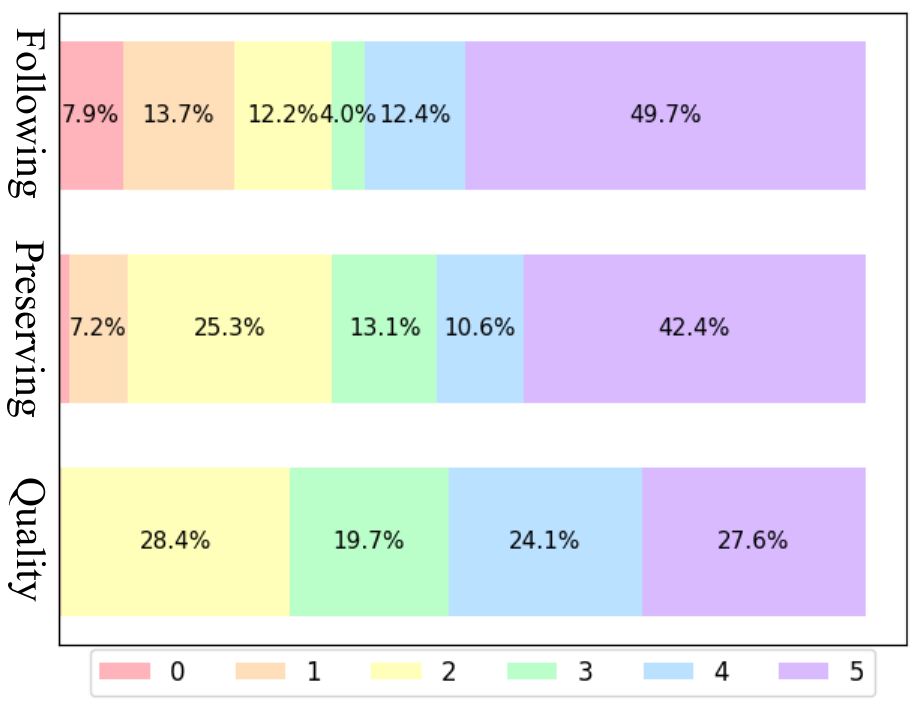}
	\caption{Distribution of reward score.}
	\label{fig:score}
    \end{minipage}%
    \begin{minipage}{0.66\textwidth}
        \includegraphics[width=1.0\linewidth]{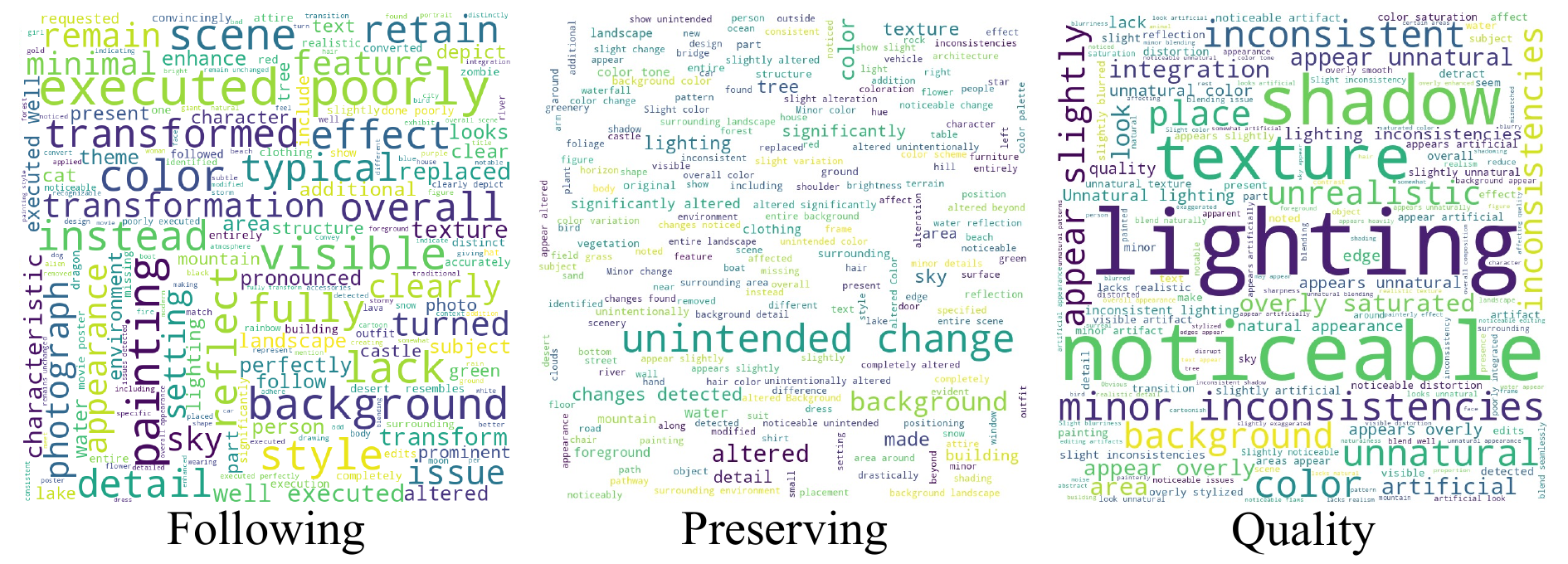}
	\caption{Word cloud of reward text.}
	\label{fig:ciyun}
    \end{minipage}
\end{figure}

\noindent
\textbf{Data statistics.} We summarize the statistics of the reward data. Fig.\ref{fig:score} shows the distribution of the reward scores, revealing that in all three aspects, samples with scores less than 5 exceed 50\%, indirectly indicating that the majority of training samples in the InsPix2Pix dataset remain noisy. Fig.\ref{fig:ciyun} uses word clouds to display the most frequent words in the reward texts. These words reflect the main issues present in the original dataset. For example, the high frequency of `executed' and `poorly' in the instruction-following aspect indicates failures in following instructions, `unintended' and `change' in the detail-preserving aspect reflect inconsistencies in non-edit areas, and `lighting' and `shadow' in the quality aspect highlight quality issues in the edited images.

\section{Methodology}

\textbf{Overview.}  
In this section, we first introduce the most general image editing framework (Sec.\ref{sec:1}). Then, we present our framework that uses multi-reward as an input condition (Sec.\ref{sec:2}). Finally, we offer a detailed explanation of the multi-reward condition module (Sec.~\ref{sec:3}).

\begin{figure}[ht]
	\centering
	\includegraphics[width=\linewidth]{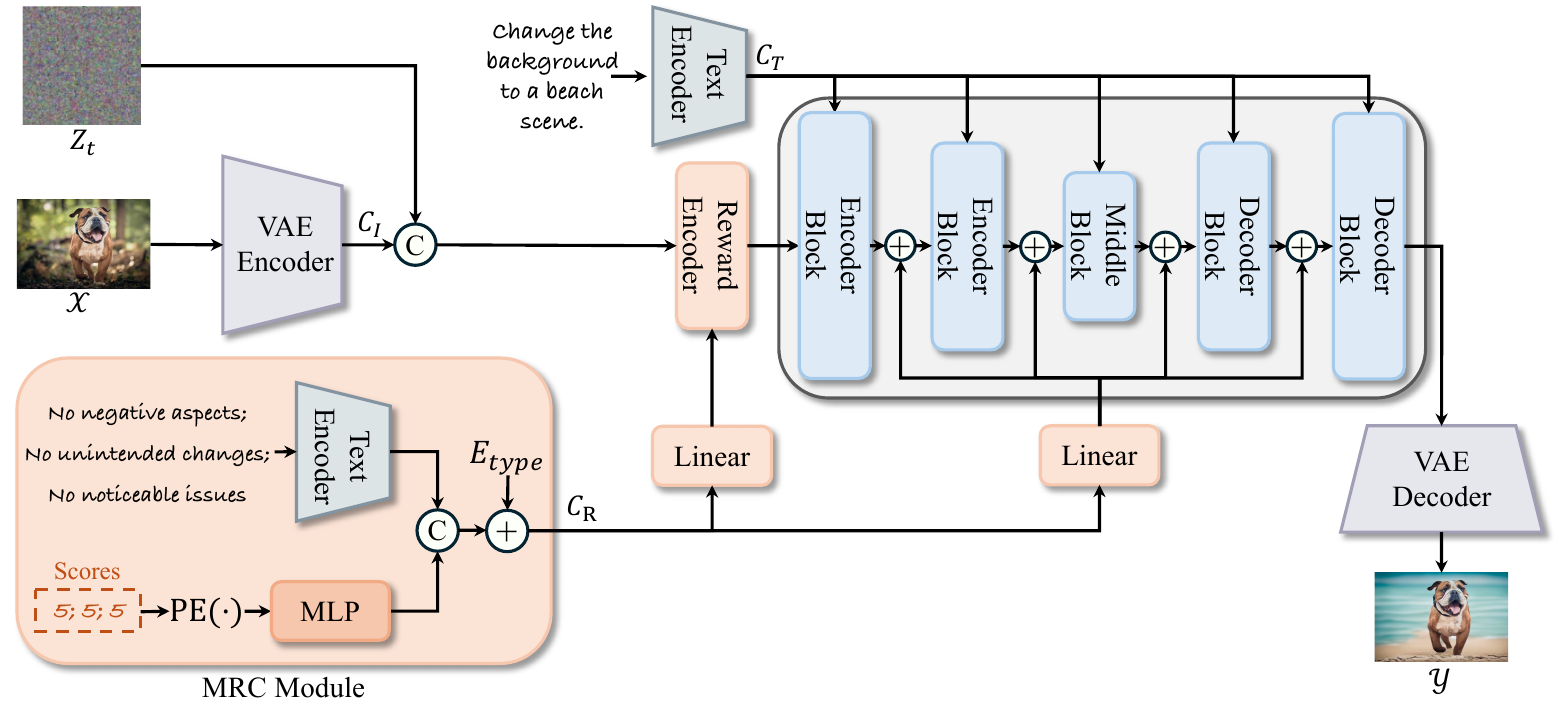}
	\caption{The overall framework of our approach. The original image $x$ is first encoded into an image condition by the VAE encoder. This image condition $c_I$ is then concatenated with latent noise $Z_t$ to serve as the query for the reward encoder, with the reward condition $c_R$ as the key/value. The resulting latent noise, containing reward information, is used as the input for the U-Net module. Meanwhile, the instruction is encoded into a text condition $c_T$ by the text encoder, which is fed into each block of the U-Net. To further enhance reward guidance, we incorporate the reward condition after each block. Finally, the U-Net's output is decoded by the VAE decoder into the edited image $y$.}
	\label{fig:network}
\end{figure}

\subsection{Preliminary: General Image Editing Framework}
\label{sec:1}

InsPix2Pix~\citep{instructpix2pix}, as one of the pioneering works in the field of instruction-based image editing, can edit images according to the given instructions. Specifically, given the original image $x$, the text instruction $t$, and the edited image $y$, first use the VAE encoder to extract the encoded latent $z$ and original image conditioning $c_I$, that is, $z = \mathcal{E}(y), c_I = \mathcal{E}(x)$. Similarly, use the text encoder to extract the text condition $c_T$. Through the diffusion process, noise is added to $z$ to generate latent noise $z_t$, where the noise level increases over timesteps $t \in T $. Then, train a network that predicts the noise added to the noisy latent $z_t$ given the original image conditioning $c_I$ and the text instruction conditioning $c_T$. The specific objective of latent diffusion is as follows: 
{
\begin{align}
\mathcal{L}_{\mathrm{InsPix2Pix}}=\mathbb{E}_{z, c_I, c_{T}, \epsilon \sim \mathcal{N}(0,1), t}[\| \epsilon -\epsilon_\delta(t, \mathrm{concat}[z_t, c_I], c_T)) \|_2^2]
\label{eq:diffusion}
\end{align}
} where $\epsilon$ is the unscaled noise, $t$ is the sampling timestep, $z_{t}$ is latent noise at step $t$. SmartEdit~\citep{smartedit}, the state-of-the-art instruction-based image editing model, uses the same architecture as InsPix2Pix but upgrades text encoder from CLIP~\citep{radford2021learning} to LLaVA~\citep{llava}.

\noindent
Although methods like InsPix2Pix and SmartEdit have shown compelling results in image editing, they are still affected by noise present in the training data, thus limiting their performance. To address this, we propose using multi-perspective rewards as an additional condition to correct the bias introduced by the training data.

\subsection{Multi-Reward as Input Condition}
\label{sec:2}

We adopted an architecture similar to InsPix2Pix and SmartEdit. On this basis, to utilize rewards to guide the model, we designed a multi-reward condition (MRC) module to extract the reward condition and used a reward encoder to integrate the reward condition into the diffusion process. Additionally, to further enhance the guidance of reward information, we also incorporated the reward condition after each block in the U-Net module. The framework as shown in Fig.~\ref{fig:network}. Given the original image $x$, the edited image $y$ is generated under the guidance of the instruction text $t$ and the reward data. 
First, we use the VAE encoder to extract the latent representation $c_I$ of the original image. As in InsPix2Pix, concatenate $c_I$ with latent noise $Z_t$ and fuse them through convolution to obtain $Z_t'$. Then, we use the proposed MRC module to generate a reward condition $c_R$ (Details in Sec.~\ref{sec:3}). To utilize the reward condition $c_R$ to guide image editing, we integrate the reward condition $c_R$ into the encoded latent noise through a reward encoder, which consists of {$1$ standard transformer encoder block~\citep{vaswani2017attention}}. Specifically, let latent noise $Z_t'$ serve as query and reward condition as key/value, this process can be expressed as follows,
{
\begin{align}
    Z_t'' = \mathtt{CA}(Z_t', \mathtt{Linear{_1}}(c_R))
\end{align}
} where $\mathtt{CA}(\textbf{z},\textbf{u})$ denotes the {transformer encoder block} with $\textbf{z}$ generating query and $\textbf{u}$ is the key/value. {$\mathtt{Linear{_1}}(\cdot)$ denotes linear projection, which aligns the dimension of the reward condition with the latent noise}. To further enhance the guidance of the reward information, we also add the reward condition after each block in the U-Net module. The input to the $i$ th block in U-Net is as follows,
{
\begin{align}
   \hat{z}_i = \mathtt{UB}_{i-1}(\hat{z}_{i-1}) + \mathtt{Linear{_2}}(c_R)
\end{align}
} where $\mathtt{UB}_{i-1}(\cdot)$ denotes $i$-1 th the blocks in U-Net. {$\mathtt{Linear{_2}}(\cdot)$ aligns the dimension of the reward condition with the U-Net}. After that, the output of the U-Net module is fed into the VAE decoder to generate the edited image $y$. The specific process can be formulated as:
{
\begin{align}
\mathcal{L}_{\mathrm{Reward}}=\mathbb{E}_{z, c_I, c_{T}, c_R , \epsilon \sim \mathcal{N}(0,1), t}[\| \epsilon -\epsilon_\delta(t, \mathrm{concat}[z_t, c_I], c_T, c_R)) \|_2^2]
\end{align}
}

\subsection{Multi-Reward Condition Module}
\label{sec:3}

We use an additional reward condition $c_R$ from the MRC module to guide the model in generating the desired edited image. The MRC module is responsible for generating the reward condition from the reward text and reward score. For the reward text, we use the text encoder in Stable Diffusion model to extract text embeddings $E_t$, as follows:
{
\begin{align}
    E_t = \mathtt{Encoder_{text}}(\mathtt{Concat}[\mathcal{T}_f, \mathcal{T}_p, \mathcal{T}_q])
\end{align}
} where $\mathcal{T}_f, \mathcal{T}_p, \mathcal{T}_q$ are the reward text in terms of following, preserving, and quality, respectively. For the reward scores, we use absolute positional encoding~\citep{vaswani2017attention}, which utilizes sine and cosine functions to convert the scores into vectors, and then extract embeddings $E_s$ with an MLP module. The process mentioned above is represented as:
{
\begin{align}
    E_s = \mathtt{MLP}(\mathtt{Concat}[\mathtt{PE}(\mathcal{S}_f), \mathtt{PE}(\mathcal{S}_p), \mathtt{PE}(\mathcal{S}_q)])
\end{align}
} where $\mathcal{S}_f, \mathcal{S}_p, \mathcal{S}_q$ are the reward scores in terms of following, preserving, and quality, respectively. $\mathtt{MLP}(\cdot)$ and $\mathtt{PE}(\cdot)$ denote the MLP module and position encoding. Finally, concatenate the text embedding and the score embedding, and add the type embedding to obtain the reward condition $c_R$.

\section{Evaluation benchmark and metrics}
\label{sec:eval}

\begin{figure}[t]
    \centering
    \begin{minipage}{0.33\textwidth}
        \centering
        \includegraphics[width=\textwidth]{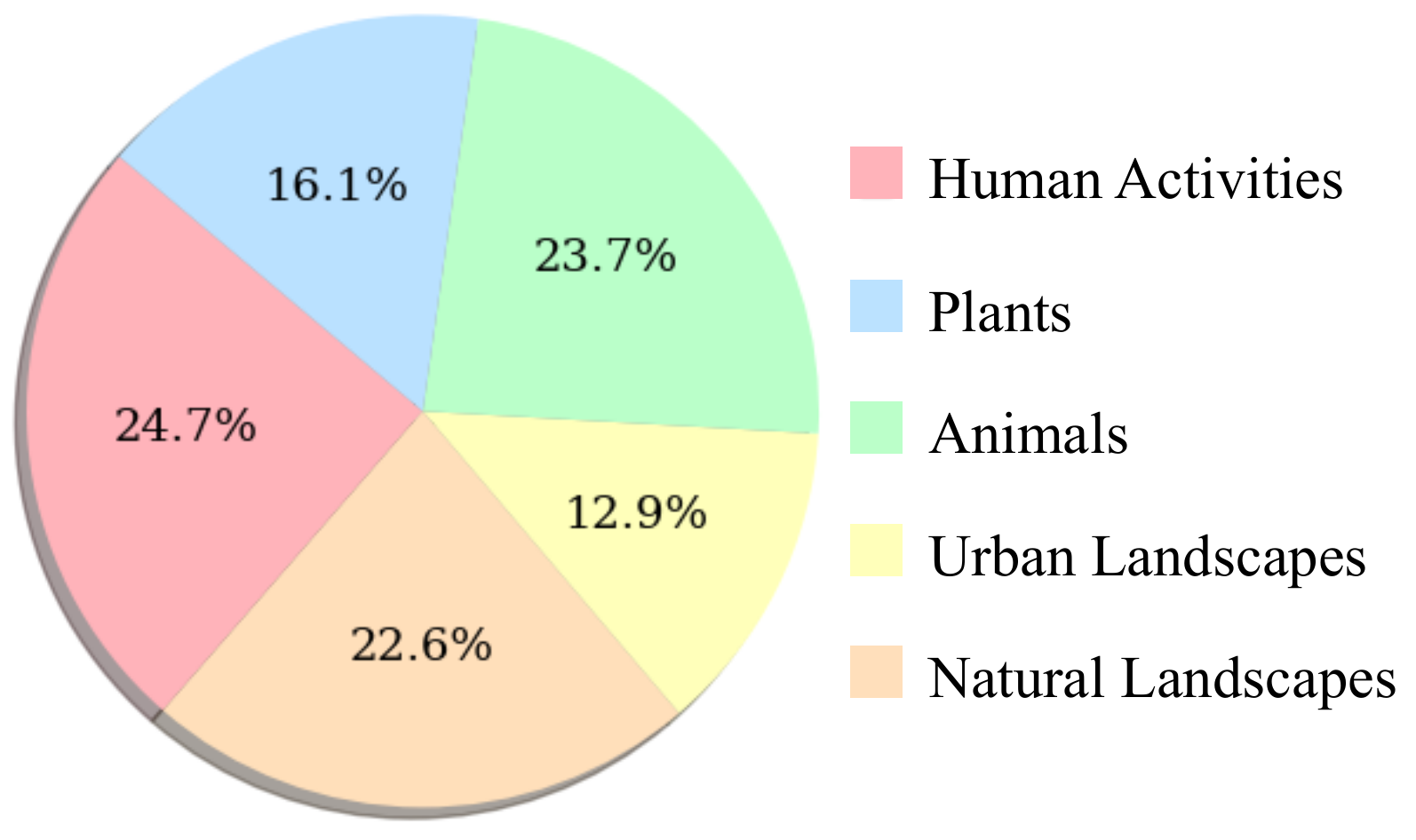}
        \caption{Distribution of different categories of images in Real-Edit.}
        \label{fig:distribution}
    \end{minipage}%
    \begin{minipage}{0.66\textwidth}
        \centering
        \includegraphics[width=\textwidth]{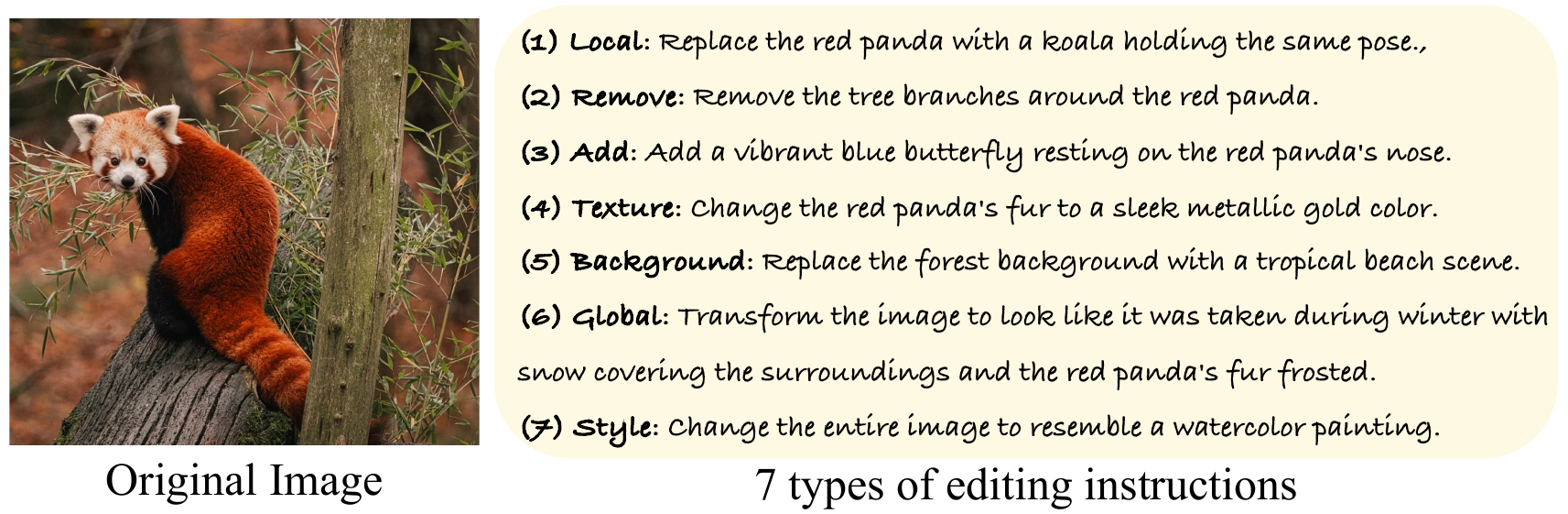}
        \caption{An example in Real-Edit.}
        \label{fig:example}
    \end{minipage}
\end{figure}

\textbf{Evaluation Data.} To more comprehensively evaluate the model's ability to edit real images based on instructions, we constructed a new image editing evaluation benchmark, Real-Edit, using real-world images. Compared to existing evaluation benchmarks, our proposed test set includes higher-quality images and a greater variety of editing instructions. We first carefully selected 80 high-quality images from the Unsplash website as the original images. The categories of these images are shown in Fig.\ref{fig:distribution}. Then, using GPT-4o, we generated 7 different editing instructions for each image based on its content, including local, remove, add, texture, background, global, and style edits, as shown in Fig.\ref{fig:example}.

\vspace{-0.4cm}

\begin{figure}[t]
    \vspace{-0.3cm}
	\centering
	\includegraphics[width=1.\linewidth]{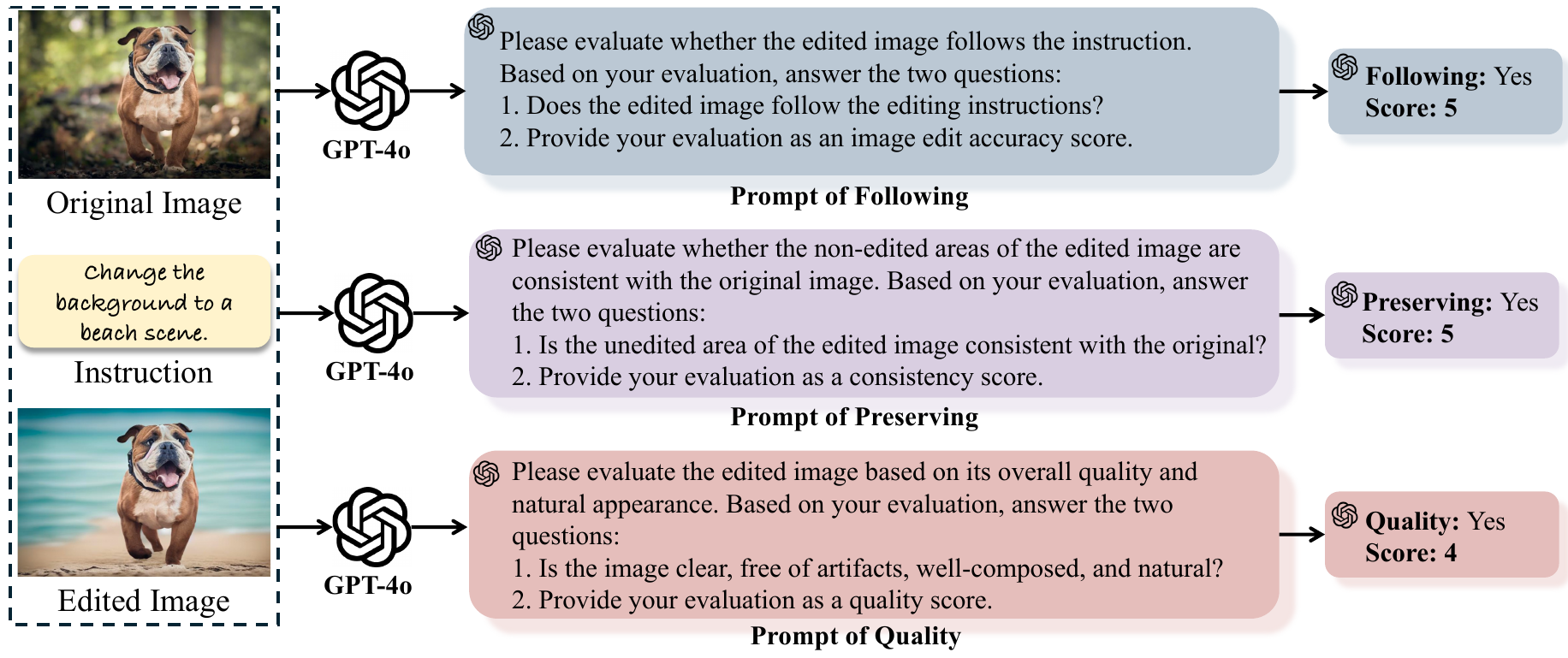}
	\caption{Evaluation process. The generated edited image, original image, and instruction are input into GPT. Three prompts are designed to evaluate from three different aspects. For each aspect, determine whether the criteria are met and assign a score (ranging from 0 to 5).}
	\label{fig:evaluate}
    \vspace{-0.5cm}
\end{figure}

\noindent\textbf{Evaluation Metrics.} To more accurately evaluate the performance of the editing model, we used GPT-4o to evaluate the edited images based on the original images and instructions. The evaluation is conducted from three perspectives as follows: (1) Following: Determine whether the edited image has been modified according to the editing instructions. (2) Preserving: Evaluate whether the non-edited aspects of the original and edited images remain consistent. (3) Quality: Focuses on the overall quality of the edited image compared to the input image, including aspects such as clarity, composition, and lighting. The detailed evaluation process is illustrated in Fig.~\ref{fig:evaluate}. Due to limited space, we only show the core prompts in the figure, while the complete prompts are provided in the appendix. 
In the later quantitative results, we have supplemented each edited image with evaluation scores.

\vspace{-0.4cm}

\section{Experiments}

\subsection{Implementation Details.}

Our method is implemented in Python using PyTorch. The MRC module, reward encoder, and the connected linear layer are randomly initialized. All other modules are initialized from the pre-trained InsPix2Pix model~\citep{instructpix2pix}. During training, we only optimize the MRC module, the U-Net module, the reward encoder, and the connected linear layers. And we use the Adam \citep{kingma2014adam} optimizer with an initial learning rate of $5e-5$, a weight decay of $1e-2$, and a warm-up ratio of $0$. We resize the images to $256$ and apply random cropping during training and resize the shorter side to $512$ during inference.

\subsection{State-of-the-art Comparison}

\begin{table}[tb]
\centering
\caption{Comparison with existing state-of-the-art methods on Real-Edit.}
\vspace{-0.2cm}
\label{tab:1}
\renewcommand{\arraystretch}{1.1}
\scalebox{0.95}{
\begin{tabular}{lccccccc}
\specialrule{1.5pt}{0pt}{0pt}
\rowcolor{mygray}
\multirow{2}{*}{} & \multirow{2}{*}{} & \multicolumn{2}{c}{Following} & \multicolumn{2}{c}{Preserving} & \multicolumn{2}{c}{Quality} \\  
\rowcolor{mygray}
\multirow{-2}{*}{\cellcolor{mygray} Method} & \multirow{-2}{*}{\cellcolor{mygray} Edit Data} & Acc & Score & Acc & Score & Acc & Score \\ \hline
KOSMOS-G~\citep{kosmosg} & 9M & 51\% & 2.82 & 9\% & 1.43 & 27\% & 3.20 \\
MagicBrush~\citep{magicbrush} & 0.31M & 51\% & 2.90 & 70\% & 3.85 & 50\% & 3.67 \\
MGIE~\citep{mgie} & 1M & 40\% & 2.43 & 45\% & 2.79 & 38\% & 3.35 \\ 
InstructDiffusion~\citep{instructdiffusion} & 0.86M  & 52\% & 2.87 & 54\% & 3.17 & 47\% & 3.58 \\
HIVE~\citep{hive} & 1.1M  & 54\% & 2.93 & 56\% & 3.36 & 53\% & 3.72 \\
HQ-Edit~\citep{hui2024hq} & 0.5M & 51\% & 2.84 & 16\% & 1.63 & \textbf{54\%} & \textbf{3.84} \\
\hline
InsPix2Pix~\citep{instructpix2pix} & 0.3M & 52\% & 2.94 & 53\% & 3.31 & 50\% & 3.69 \\
Reward-InsPix2Pix & 0.32M & 63\% & 3.39 & 58\% & 3.43 & \textbf{54\%} & 3.80 \\ \hline
SmartEdit~\citep{smartedit} & 1.17M & 64\% & 3.50 & 66\% & 3.70 & 45\% & 3.56\\
Reward-SmartEdit & 1.19M & \textbf{69\%} & \textbf{3.72} & \textbf{74\%} & \textbf{4.00} & 49\% & 3.67 \\ 
\specialrule{1.5pt}{0pt}{0pt}
\end{tabular}}
\end{table}

\begin{table}[!htbp]
    \centering
    \begin{minipage}{0.47\textwidth}
    \vspace{-0.2cm}
    \small
        \caption{Results on MagicBrush test set (\%).}
        \vspace{-0.2cm}
        \label{tab:2}
        \renewcommand{\arraystretch}{1.1}
        \begin{tabular}{lcc}
        \specialrule{1.5pt}{0pt}{0pt} 
        \rowcolor{mygray}
        Method & \small{CLIP-I} & \small{CLIP-T}  \\ \hline
        MagicBrush & 90.7 & \textbf{30.6} \\
        MGIE & 90.9 & 30.5 \\ 
        InstructDiffusion & 89.2 & 30.2 \\ \hline
        InsPix2Pix & 85.4 & 29.2 \\
        Reward-InsPix2Pix & 88.9 & 29.8 \\ \hline
        SmartEdit  & 90.4 & 30.3 \\ 
        Reward-SmartEdit & \textbf{91.3} & 30.5  \\ 
        \specialrule{1.5pt}{0pt}{0pt}
        \end{tabular}
    \end{minipage}%
    \begin{minipage}{0.53\textwidth}
        \centering
        \small
        \caption{Human evaluation scores for SmartEdit and Reward-SmartEdit on Real-Edit benchmark.}
        \label{tab:human}
        \renewcommand{\arraystretch}{1.1}
        \begin{tabular}{cccc}
        \specialrule{1.5pt}{0pt}{0pt}
        \rowcolor{mygray}
        Method & \small{Following} & \small{Preserving} & \small{Quality} \\ \hline
        SmartEdit & 3.09 & 3.18 & 2.58 \\
        Reward-SmartEdit & \textbf{3.34} & \textbf{3.46} & \textbf{2.73} \\
        \specialrule{1.5pt}{0pt}{0pt}
        \end{tabular}
    \end{minipage}
\vspace{-0.2cm}
\end{table}

To validate the efficacy of our method, we compared it against other image editing methods. The results on Real-Edit are summarized in Tab.~\ref{tab:1}. Reward-InsPix2Pix, which is fine-tuned based on reward data, significantly improved all metrics compared to InsPix2Pix: following accuracy increased by 11\%, score by 0.45, preserving accuracy by 5\%, score by 0.12, quality accuracy by 4\%, and score by 0.11. SmartEdit, as the leading image editing model, achieved a new SOTA performance after fine-tuning based on reward data. 
{The proposed MRC module needs to be trained separately for each model.}
Despite using much less additional editing data (0.02M) compared to KOSMOS-G (9M), our method significantly improved both InsPix2Pix and SmartEdit, demonstrating its efficiency.

\noindent
We also evaluated our method on the common evaluation benchmark MagicBrush~\citep{magicbrush}, as shown in Tab.~\ref{tab:2}. Using reward data for fine-tuning still improves the performance of the editing model. Specifically, it helps InsPix2Pix improve by 3.5 on CLIP-I and 0.6 on CLIP-T, and it helps SmartEdit improve by 0.9 on CLIP-I and 0.2 on CLIP-T.  These results once again validate the effectiveness of our method.

\subsection{Human Evaluation}

To further validate the performance of our method against state-of-the-art methods, we conduct a human evaluation. Specifically, we selected the best-performing SmartEdit and the Reward-SmartEdit model fine-tuned using reward data. We collected the edited images they generated on Real-Edit, with 560 samples each. We then recruited 10 professional annotators to evaluate the edited images based on the three aforementioned aspects. The evaluation results are shown in Tab.~\ref{tab:human}. As indicated in the table, Reward-SmartEdit significantly outperformed the original SmartEdit, further demonstrating the effectiveness of our method. The human evaluation score is in general lower than GPT-4o scores on all methods (Fig. \ref{fig:human} in Appendix), but the rank of different methods are consistent.
We guess that the reason for this discrepancy may be that human evaluators often have higher expectations and subjective perceptions, making them more critical of details and quality.

\subsection{Ablation Study}

\textbf{Ablation for two types of reward data.} The reward data consists of reward scores and reward text. To explore the effects of these two types of reward information, we conducted ablation experiments on Real-Edit. As shown in Tab.~\ref{tab:3}, our baseline model without reward information (\ding{182}) achieves a following accuracy of 49\%, preserving the accuracy of 38\%, and quality accuracy of 32\%. When the reward score is applied alone, these metrics improve by 12\%, 17\%, and 21\%, respectively (\ding{182} \emph{vs.} \ding{183}). When the reward text is used alone, the metrics improve by 11\%, 14\%, and 19\%, respectively (\ding{182} \emph{vs.} \ding{184}). Combining both reward score and reward text yields the best results, with the following accuracy, preserving accuracy, and quality accuracy reaching 63\%, 58\%, and 54\%, respectively (\ding{185}). These results clearly validate the efficacy of incorporating reward information.

\begin{table}[htb]
\centering
\caption{Ablation of two types of reward data on Real-Edit.}
\label{tab:3}
\renewcommand{\arraystretch}{1.1}
\begin{tabular}{ccc|cccccc}
\specialrule{1.5pt}{0pt}{0pt}
\rowcolor{mygray}
& \multicolumn{2}{c|}{Reward} & \multicolumn{2}{c}{Following} & \multicolumn{2}{c}{Preserving} & \multicolumn{2}{c}{Quality} \\  
\rowcolor{mygray}
& Score & Text & Acc & Score & Acc & Score & Acc & Score \\ \hline
\ding{182} & & & 49\%  & 2.77  & 38\% & 2.59  & 32\% & 3.15 \\
\ding{183} & \checkmark& & 61\%  & 3.29  & 55\% & 3.40 & 53\%  & 3.70 \\
\ding{184}& & \checkmark & 60\% & 3.25 & 52\% & 3.30  & 51\% & 3.25  \\
\ding{185}& \checkmark & \checkmark & \textbf{63\%} & \textbf{3.39} & \textbf{58\%} & \textbf{3.43} & \textbf{54\%} & \textbf{3.80} \\ 
\specialrule{1.5pt}{0pt}{0pt}
\end{tabular}
\end{table}

\noindent
\textbf{Ablation for different methods of reward integration.} To utilize reward information to guide the image editing model, we integrate the reward condition into the edit model using two methods: attention in the reward encoder and addition in the U-Net. To explore the impact of these methods, we conducted ablation experiments, as shown in Tab.~\ref{tab:4}. When using attention alone, following, preserving, and quality accuracy improved by 11\%, 8\%, and 16\%, respectively (\ding{182} \emph{vs.} \ding{183}). Using addition alone, the metrics improved by 8\%, 18\%, and 20\% (\ding{182} \emph{vs.} \ding{184}). Combining both attention and addition achieved the best performance across all metrics (\ding{185}), validating the importance of these methods for integrating reward conditions.

\begin{table}[htb]
\centering
\caption{Ablation of methods for integrating reward information on Real-Edit.}
\label{tab:4}
\renewcommand{\arraystretch}{1.1}
\begin{tabular}{ccc|cccccc}
\specialrule{1.5pt}{0pt}{0pt}
\rowcolor{mygray}
& \multirow{2}{*}{} & \multirow{2}{*}{} & \multicolumn{2}{c}{Following} & \multicolumn{2}{c}{Preserving} & \multicolumn{2}{c}{Quality} \\  
\rowcolor{mygray}
& \multirow{-2}{*}{\cellcolor{mygray} Attention} & \multirow{-2}{*}{\cellcolor{mygray} Addition} & Acc & Score & Acc & Score & Acc & Score \\ \hline
\ding{182} & & & 49\%  & 2.77  & 38\% & 2.59  & 32\% & 3.15\\
\ding{183} &\checkmark&  & 60\% & 3.27 & 46\% & 3.11 & 48\% & 3.64  \\
\ding{184} && \checkmark & 57\% & 3.15 & 56\% & \textbf{3.44} & 52\% & 3.76  \\
\ding{185} & \checkmark & \checkmark &  \textbf{63\%} & \textbf{3.39} & \textbf{58\%} & 3.43 & \textbf{54\%} & \textbf{3.80}  \\ 
\specialrule{1.5pt}{0pt}{0pt}
\end{tabular}
\end{table}

\noindent
\textbf{Ablation for different reward scores during inference.} During inference, we set the reward scores for following, preserving, and quality to 5, and set the reward text to `None'. To investigate whether the model's editing performance is influenced by reward information, we conducted experiments as shown in Tab.\ref{tab:score}. From the results in Tab.\ref{tab:score}, we observe that as the scores decrease, the model's editing performance in all three aspects significantly declines. Specifically, when the scores dropped from 5 to 0, the accuracy for following, preserving, and quality decreased by 9\%, 28\%, and 19\%, respectively. This indicates that our model can understand the meanings of different scores and achieve a certain degree of controllable generation in the quality of the generated images.

\begin{table}[htb]
\centering
\caption{Ablation of different reward scores.}
\label{tab:score}
\renewcommand{\arraystretch}{1.1}
\begin{tabular}{cccc|cccccc}
\specialrule{1.5pt}{0pt}{0pt}
\rowcolor{mygray}
& \multicolumn{3}{c|}{Reward Score} & \multicolumn{2}{c}{Following} & \multicolumn{2}{c}{Preserving} & \multicolumn{2}{c}{Quality} \\  
\rowcolor{mygray}
&F & P & Q & Acc & Score & Acc & Score & Acc & Score \\ \hline
\ding{182} & 0 & 0 & 0 & 54\% & 2.90 & 30\% & 2.31 & 35\% & 3.37 \\
\ding{183} & 3 & 3 & 3 & 59\% & 3.19 & 42\% & 2.94 & 47\% & 3.64 \\
\ding{184} & 5 & 5 & 5 &  \textbf{63\%} & \textbf{3.39} & \textbf{58\%} & \textbf{3.43} & \textbf{54\%} & \textbf{3.80}   \\
\specialrule{1.5pt}{0pt}{0pt}
\end{tabular}
\end{table}

\vspace{-0.5cm}
\subsection{Qualitative Comparison}

\begin{figure}[ht]
	\centering
	\includegraphics[width=1.0\linewidth]{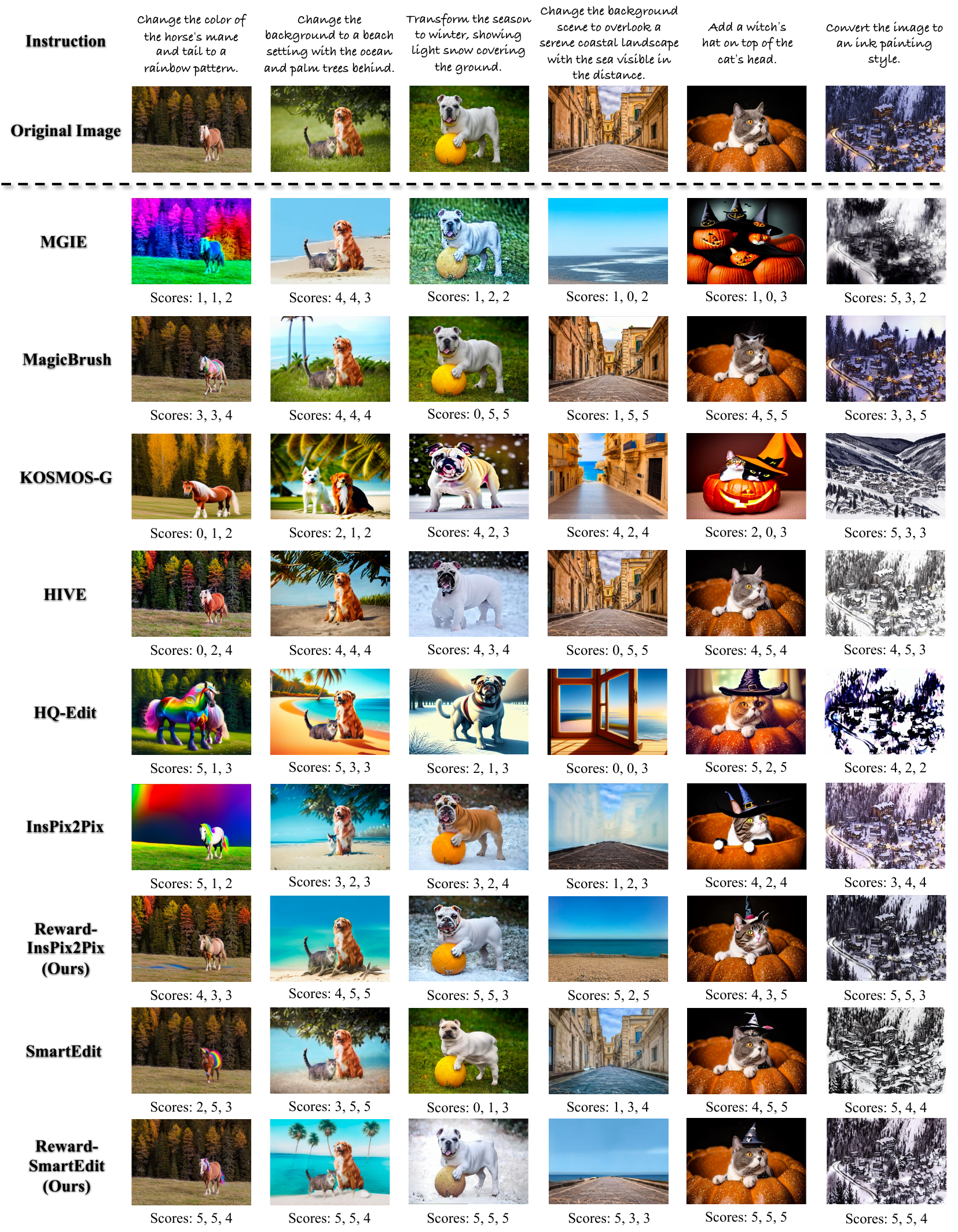}
	\caption{Qualitative results on Real-Edit. The three scores below the image are given by GPT-4o in three aspects: instruction following, detail preservation, and generation quality.}
	\label{fig:examples}
 \vspace{-0.2cm}
\end{figure}

To further qualitatively validate the effectiveness of our proposed method, we presented the results of our method on InsPix2Pix and SmartEdit, as well as the results of other image editing methods, as shown in Fig.~\ref{fig:examples}. In Fig.~\ref{fig:examples}, both the Reward-InsPix2Pix and Reward-SmartEdit outperform the original InsPix2Pix and SmartEdit, and their editing performance is also better compared to other methods, showing the effectiveness of our method.

\section{Conclusion}

We propose a novel framework to rectify the noisy supervision for instruction-based image editing models by adding multi-perspective reward data as additional conditions. We collect 20k multi-perspective reward data, named RewardEdit-20k, using a subset of InxPix2Pix dataset and GPT-4o. Additionally, we presented the Real-Edit benchmark and a GPT-4o-based evaluation method. Extensive experiments show that our approach significantly enhances performance across all perspectives, achieving state-of-the-art results in both GPT-4o and human evaluations. 

{\small
\bibliography{iclr2025_conference}
\bibliographystyle{iclr2025_conference}
}
\newpage
\appendix

\section*{\textbf{Appendix}}

\section{Failure Case Analysis}
\begin{figure}[ht]
	\centering
	\includegraphics[width=1.\linewidth]{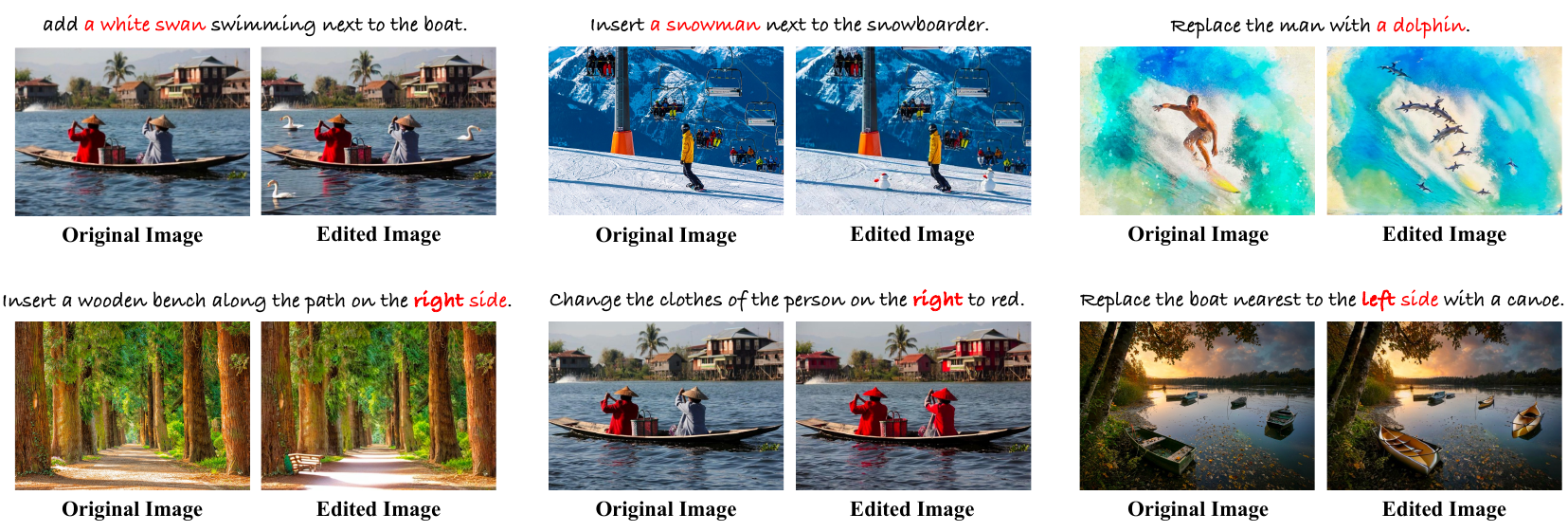}
	\caption{Examples of failure cases. In the first three edited images, the number of objects is incorrect, while in the last three edited images, the spatial positions of the objects are incorrect.}
	\label{fig:failure}
\end{figure}

To explore the limitations of our method, we collected and analyzed failed cases. The analysis revealed two main limitations of our method. 
The first limitation is that during testing, even when the given multi-perspective reward scores are all 5, the generated edited image does not always achieve a score of 5. This indicates that the reward information does not always perfectly guide the model, especially in some complex cases. The second limitation is that our method has difficulty accurately understanding the quantifiers and spatial position words in the instructions, as shown in Fig.~\ref{fig:failure}. This may be due to the model's insufficient understanding of fine-grained textual features. In future work, we will explore ways to improve the model's understanding of fine-grained semantics for image editing.

\section{Examples in RewardEdit-20K}
\label{secb}

\begin{figure}[htb]
	\centering
	\includegraphics[width=1.\linewidth]{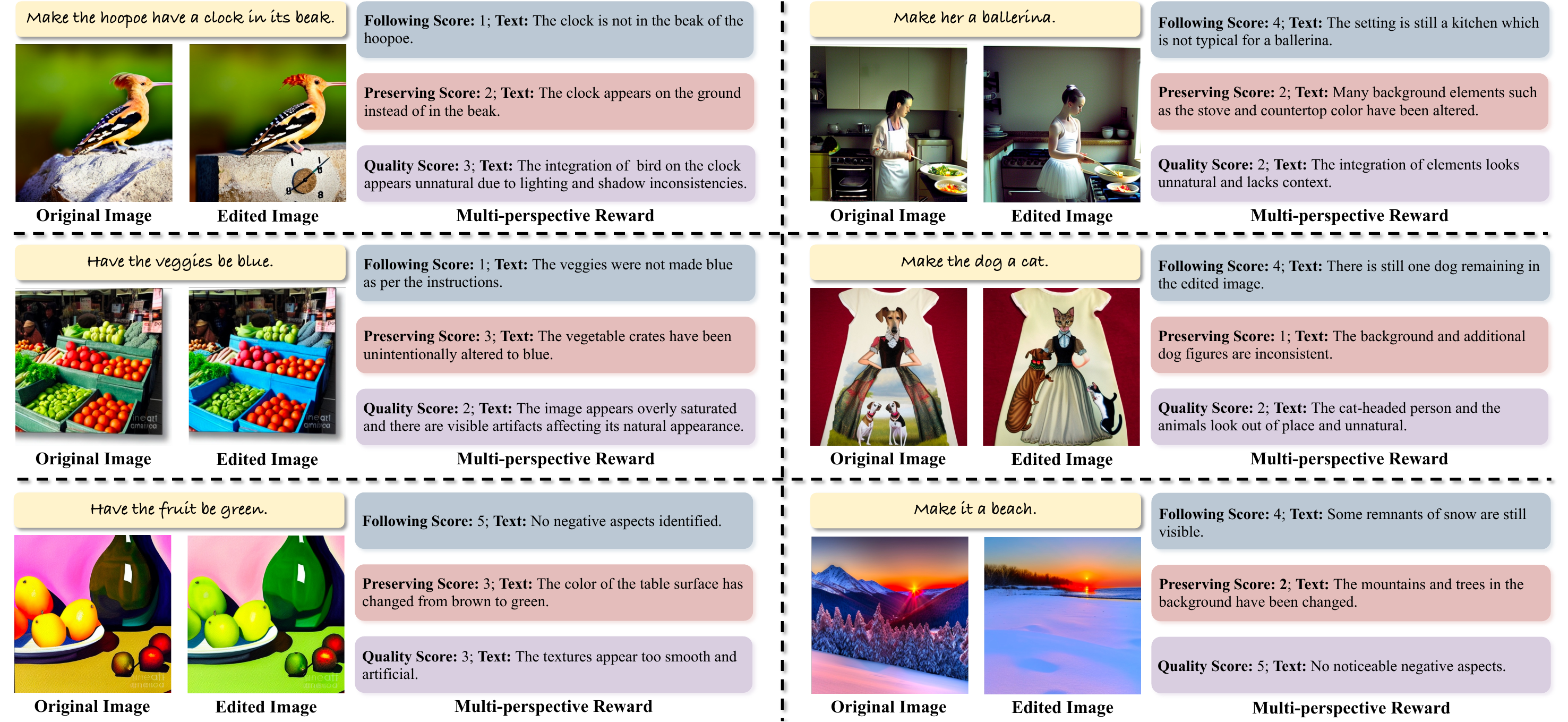}
	\caption{Examples from the RewardEdit-20K dataset. Best viewed with zoom-in.}
	\label{fig:ex}
\end{figure}

We show examples from RewardEdit-20K, as shown in Fig.~\ref{fig:ex}. For each triplet (instruction, original image, edited image), there are three perspectives of rewards: instruction following, detail preserving, and generation quality. Each reward consists of a score and text. The reward score reflects the overall quality, while the reward text provides more detailed information.

\section{{ADDITIONAL Experiment Results}}

\subsection{{Evaluation based on existing metrics}}
{We also evaluated Following, Preserving, and Quality based on existing evaluation metrics, as shown in Fig.~\ref{tab:f1}. We recalculated the performance of existing methods and our method based on the CLIP score and the FID score. Specifically, the CLIP feature similarity between the edited image and the instruction is the Following score, the similarity between the original and edited images is the Preserving score, and the FID between the original and edited images is the Quality score. The table shows that our method still achieved promising results and improvements over the baseline.
However, these metrics also have limitations: 1) When the editing instruction and the images are complicated, CLIP/FID score can not accurately represent the following/preserving/quality of the edited image, e.g., CLIP can not distinguish left/right. 2) the range of the following score and preserving score is relatively small, which may make it hard to distinguish performance differences between methods.}

\begin{table}[htb]
\centering
\caption{{Comparison of different methods based on existing evaluation metrics.}}
\label{tab:f1}
\renewcommand{\arraystretch}{1.1}
\begin{tabular}{lcccc}
\specialrule{1.5pt}{0pt}{0pt}
\rowcolor{mygray}
Method  & Following (CLIP) & Preserving (CLIP) & Quality (FID) $\downarrow$ \\ \hline
KOSMOS-G  & 26.8 & 86.4 & 3.01 \\
MagicBrush   & 25.2 & 91.9 & 2.86 \\
MGIE  & 26.4 & 87.0 & 3.09 \\
InstructDiffusion  & 26.3 & 86.4 & 2.89 \\
HIVE  & 26.3 & 89.0 & 3.08 \\
HQ-Edit  & 28.5 & 77.2 & 3.59 \\ \hline
InsPix2Pix  & 27.0 & 82.3 & 3.51 \\
Reward-InsPix2Pix  & 27.5 & 83.8 & 3.31 \\ \hline
SmartEdit  & 26.5 & 87.7 & 2.80 \\
Reward-SmartEdit & 26.9 & 90.0 & 2.77 \\
\specialrule{1.5pt}{0pt}{0pt}
\end{tabular}
\end{table}

\subsection{{Ablation study of editing data}}

{The 20K samples in our RewardEdit-20K dataset are randomly sampled from InsPix2Pix. Our motivation is that constructing a perfect image editing dataset is challenging, and the ground truth in existing image editing datasets often contains issues. Therefore, we propose using multi-perspective rewards to rectify the inaccurate supervision. To more fairly demonstrate the role of multi-perspective rewards, we conducted the ablation experiments shown in Tab.~\ref{tab:f2}. The experimental results indicate that, with the same data, using multi-perspective rewards significantly improves performance compared to the baseline, demonstrating the effectiveness of multi-perspective rewards.}

\begin{table}[htb]
\centering
\caption{{Ablation study of editing data.}}
\label{tab:f2}
\renewcommand{\arraystretch}{1.1}
\begin{tabular}{ccccc}
\specialrule{1.5pt}{0pt}{0pt}
\rowcolor{mygray}
Method & Edit Data & Following & Preserving & Quality \\ \hline
\multirow{2}{*}{Baseline} & 0.30M & 2.77 & 2.59 & 3.15 \\
 & 0.32M & 2.90 & 2.88 & 3.52 \\
Ours & 0.32M & 3.39 & 3.43 & 3.80 \\
\specialrule{1.5pt}{0pt}{0pt}
\end{tabular}
\end{table}

\subsection{{Ablation Study on Challenging Samples}}

{To investigate whether our reward model can generate better edited images for challenging editing samples in InsPix2Pix, we first randomly selected 500 samples from RewardEdit-20K with scores not exceeding 2. Then, we used our reward model to generate edited images based on the original images and instructions of these samples, and scored them using GPT-4o. The experimental results are shown in Tab.~\ref{tab:f3}. "Original" represents the average scores of the original edited images of these samples across three metrics, while "Ours" represents the scores of the edited images generated by our reward model. From the table, it can be observed that the edited images generated by our method significantly outperform the original edited images on all three metrics, indicating that our method can generate better results for these difficult cases.}

\begin{table}[htb]
\centering
\caption{{Comparison of edited images for challenging samples in InsPix2Pix.}}
\label{tab:f3}
\renewcommand{\arraystretch}{1.1}
\begin{tabular}{cccc}
\specialrule{1.5pt}{0pt}{0pt}
\rowcolor{mygray}
Method & Following & Preserving & Quality \\ \hline
Original & 1.15 & 1.69 & 1.99 \\
Ours & 2.92 & 4.10 & 3.68 \\
\specialrule{1.5pt}{0pt}{0pt}
\end{tabular}
\end{table}

\subsection{{Ablation study of each perspective reward}}
{
We find that analyzing the impact of each perspective reward is beneficial, and we conduct additional experiments by training on each perspective separately. As shown in Tab.~\ref{tab:f4}, the following score reached 3.40 with only the instruction following reward, the preserving score reached 3.54 with only the detail preserving reward, and the quality score reached 3.95 with only the generation quality reward. These results demonstrate the effectiveness of each perspective reward.}

\begin{table}[htb]
\centering
\caption{{Ablation of each perspective reward. `IF', `DP' and `GQ' are instruction following, detail preserving and generation quality reward.}}
\label{tab:f4}
\renewcommand{\arraystretch}{1.1}
\begin{tabular}{cccccc}
\specialrule{1.5pt}{0pt}{0pt}
\rowcolor{mygray}
IF & DP & GQ & Following & Preserving & Quality \\ \hline
\checkmark  &  & & 3.40 & 3.25 & 3.72 \\
 & \checkmark & & 3.23 & 3.54 & 4.00 \\
  &  & \checkmark & 3.20 & 3.23 & 3.95 \\
\checkmark  & \checkmark & \checkmark & 3.39 & 3.43 & 3.80 \\
\specialrule{1.5pt}{0pt}{0pt}
\end{tabular}
\end{table}

\subsection{Ablation study of training resolution}

We chose to train at a resolution of 256 to maintain consistency with other methods (InsPix2Pix~\citep{instructpix2pix}, SmartEdit~\citep{smartedit} and MGIE~\citep{mgie} are both trained on 256), ensuring a fair comparison. Increasing the training resolution from 256 to 512 requires about 4 times computation, so it is hard to keep the mini-batch size per GPU unchanged. Due to limited computation, we are not able to tune the hyperparameters for 512 resolution. We use gradient accumulation to keep the overall batch size and all the other hyperparameters unchanged. As shown in~Tab.~\ref{resolution}, we did not observe performance improvement compared to 256 resolution.

\begin{table}[htb]
\centering
\caption{Ablations of training image resolution.}
\label{tab:human}
\renewcommand{\arraystretch}{1.1}
\begin{tabular}{cccc}
\specialrule{1.5pt}{0pt}{0pt}
\rowcolor{mygray}
Resolution &  \small{Following} & \small{Preserving} & \small{Quality} \\ \hline
512 & 3.28 & 3.20 & 3.61 \\
256 & 3.39 & 3.43 & 3.80 \\
\specialrule{1.5pt}{0pt}{0pt}
\end{tabular}
\label{resolution}
\end{table}

\section{{Additional analysis and discussion}}

\subsection{{Role of reward text}}
{
We introduced additional reward information because the ground truth in existing image editing datasets is inaccurate (see lines 92-104). To rectify these inaccuracies, we incorporated reward scores and text (examples in Sec.~\ref{secb}). The reward score is a quantitative evaluation that reflects the overall quality. Since the same reward score can correspond to different types of errors, we further included reward text, which provides more detailed error information. Specifically, the negative text introduced can be seen as a correction to the ground truth, which means that the original ground truth plus the negative text forms the true ground truth. To ensure that the negative text serves as a guide, we integrate it into the diffusion process as an additional condition.}

\subsection{{Limitations of REWARDEDIT-20K}}
{
We proposed REWARDEDIT-20K based on Ins-Pix2Pix. Currently, most image editing models use the Ins-Pix2Pix dataset for training, including the Instructdiffusion~\citep{instructdiffusion} and SEED-Data-Edit~\citep{ge2024seed}. Ins-Pix2Pix has become the most widely used dataset in the image editing field. Recent methods, such as SmartEdit, Instructdiffusion, and SEED-Data-Edit, typically use multiple editing datasets for mixed training. Our improvements in SmartEdit demonstrate that our method is also effective for models trained with mixed datasets. In the future, we will apply the proposed reward data generation method to other datasets to see whether it brings further improvement.}

\subsection{{Reliability of GPT-4o}}
{
The annotation/evaluation from GPT-4o is not as good as human annotation. However, human annotation is very expensive and time-consuming, making it unsuitable for large-scale data generation. In contrast, GPT-4o-based data generation is scalable with reasonable quality. Moreover, our experiments demonstrate that using multi-view rewards generated by GPT-4o can still significantly improve the model's image editing performance, indicating the reliability of our method.
\noindent
The version of GPT-4o we used is `2024-08-06'. After multiple (5 times) tests, we found that the fluctuations in the accuracy of the three metrics are within 1\%, and the score fluctuations are within 0.05. This demonstrates the stability of GPT-4o.In the feature, we will also explore fine-tuning a specialized evaluation model based on existing open-source MLLMs.}

\subsection{{Comparison with DPO-Diffusion}}
{
Both DPO-Diffusion and our proposed Multi-Reward approach fundamentally aim to optimize the quality of generated images through feedback mechanisms. The main differences between our Multi-Reward and DPO-Diffusion are as follows: 1) Granularity of feedback. DPO-Diffusion's preference feedback is expressed as relative preferences, such as `Image A is better than image B',  therefore the feedback signal only has two possible states. In contrast, our Multi-Reward uses absolute numerical values and detailed text description for feedback signals (For examples, see Appendix Section B.). 2) Applicability of feedback. DPO-Diffusion is only applicable to situations with a single feedback value, whereas our approach can simultaneously incorporate multi-perspective feedback information, including instruction following, detail preserving and generation quality. 3) Training stability. We directly use feedback information as an additional condition while still employing the original Diffusion Loss. This approach is simple and effective, avoiding the training instability that DPO can introduce to the diffusion model.}

\section{More Qualitative Examples on Real-Edit}

\begin{figure}[ht]
	\centering
	\includegraphics[width=1.\linewidth]{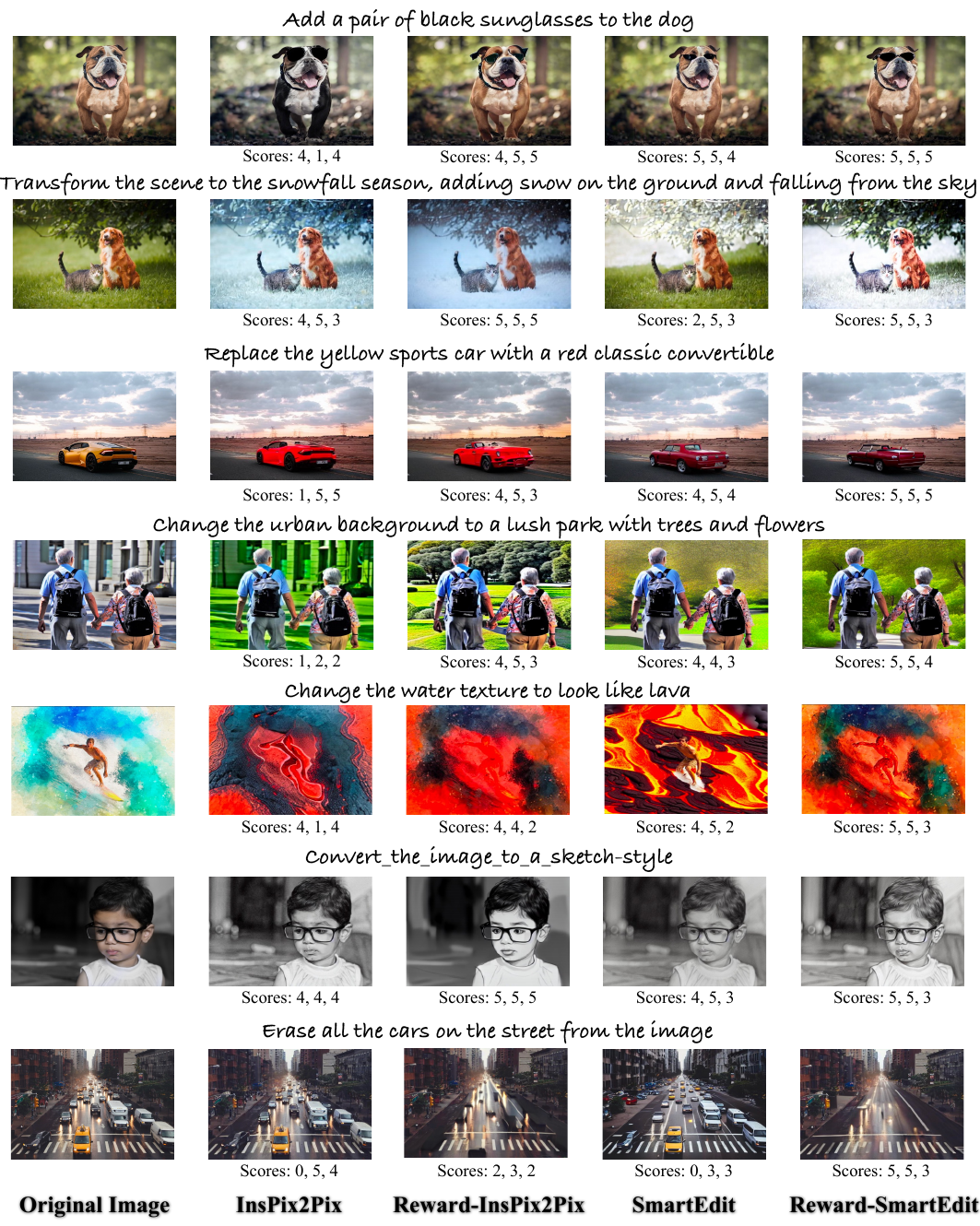}
	\caption{More quantification results on Real-Edit. The scores below the edited images are the evaluation scores given by GPT-4o.}
	\label{fig:qua}
\end{figure}

We show more visualizations of the examples, as shown in~\ref{fig:qua}. From this figure, we find that the reward-guided models, Reward-InsPix2Pix and Reward-SmartEdit, both perform better than the models without reward guidance. This further demonstrates the effectiveness of our method.

\section{Human Evaluation}

\subsection{Human Evaluation Examples}

\begin{figure}[ht]
	\centering
	\includegraphics[width=1.\linewidth]{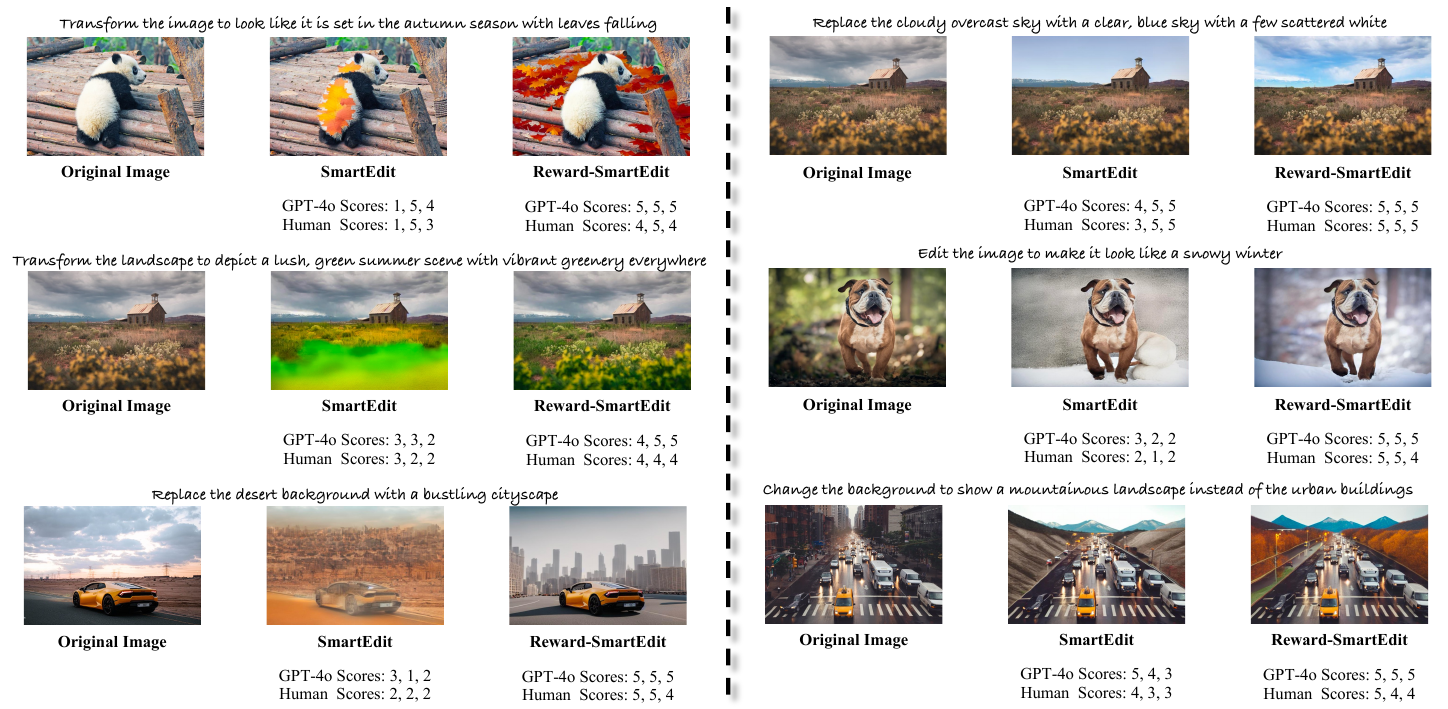}
	\caption{Examples comparing human evaluation and GPT-4o scores.}
	\label{fig:human}
\end{figure}

Fig.~\ref{fig:human} shows the scores given by GPT-4o and humans. It can be seen that the human evaluation scores and the GPT-4o scores for edited images are generally quite similar, although the human evaluation scores are overall slightly lower than the GPT-4o scores. However, both tend to give higher scores to good images and lower scores to poor images.

\subsection{Human Evaluation Interface}

\begin{figure}[ht]
	\centering
	\includegraphics[width=1.\linewidth]{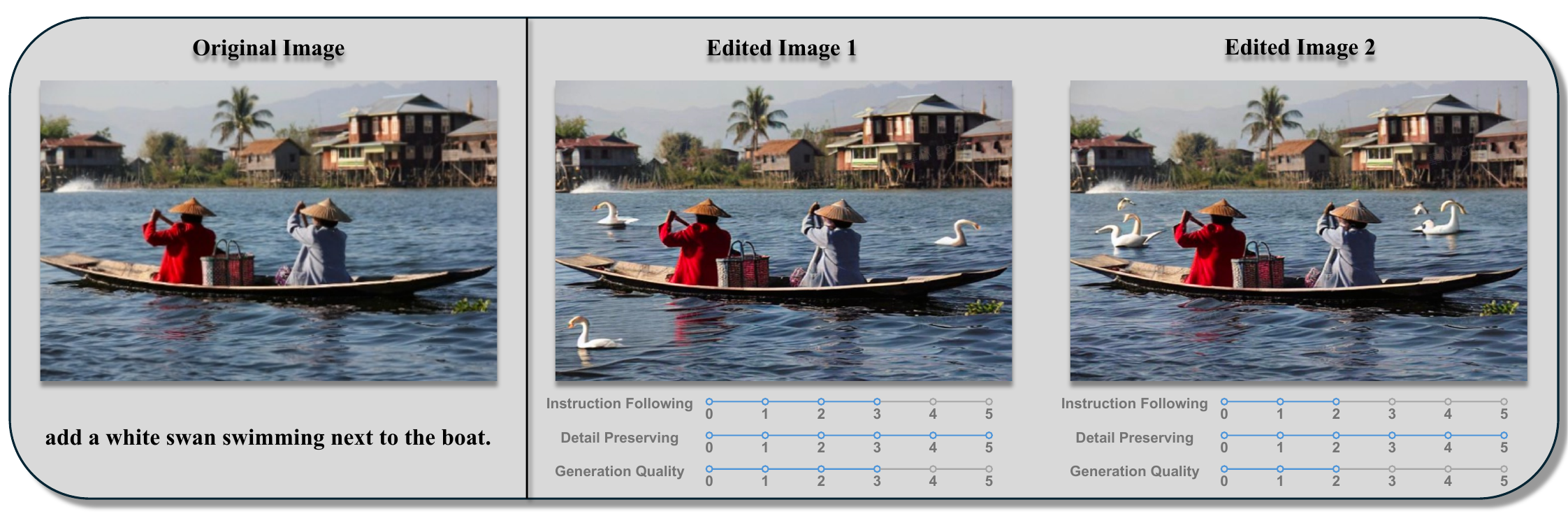}
	\caption{Human evaluation interface. Given the original image and instruction, as well as the edited images generated by SmartEdit and Reward-SmartEdit, annotators evaluate and score from three aspects.}
	\label{fig:interface}
\end{figure}

To further validate the performance of our method against state-of-the-art methods, we conducted a human evaluation. The interface is shown in Fig.~\ref{fig:interface}. The orders of "Edited Image 1" and "Edited Image 2" are randomly shuffled so that the evaluation is fair to the two methods.

\section{Complete Prompts When Using GPT-4o}

\subsection{Generate reward data}

\begin{figure}[ht]
	\centering
	\includegraphics[width=1.\linewidth]{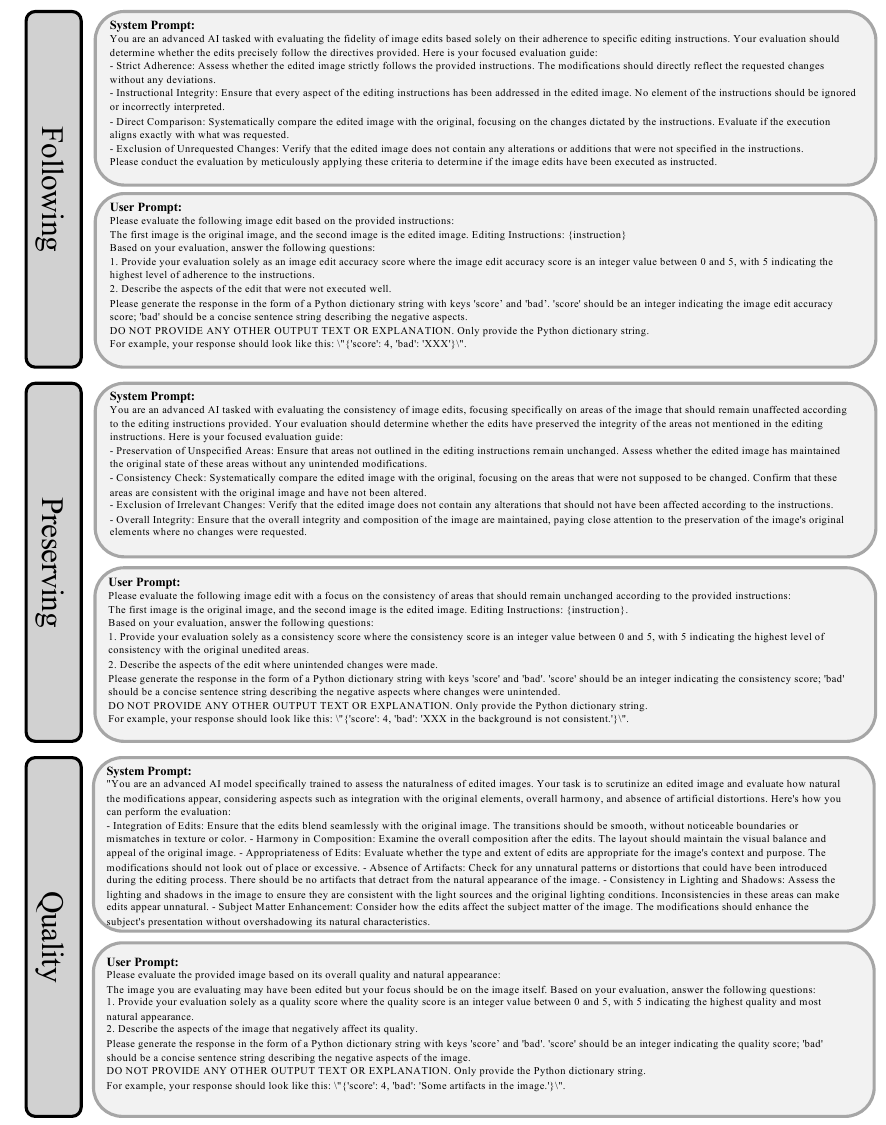}
	\caption{Complete prompts for generating reward data. Best viewed with zoom-in.}
	\label{fig:p1}
\end{figure}

We used GPT-4o to generate the multi-reward dataset RewardEdit-20K, designing three types of prompts for following, preserving, and quality. The complete prompts are shown in Fig.~\ref{fig:p1}.

\subsection{Evaluation}

\begin{figure}[ht]
	\centering
	\includegraphics[width=1.\linewidth]{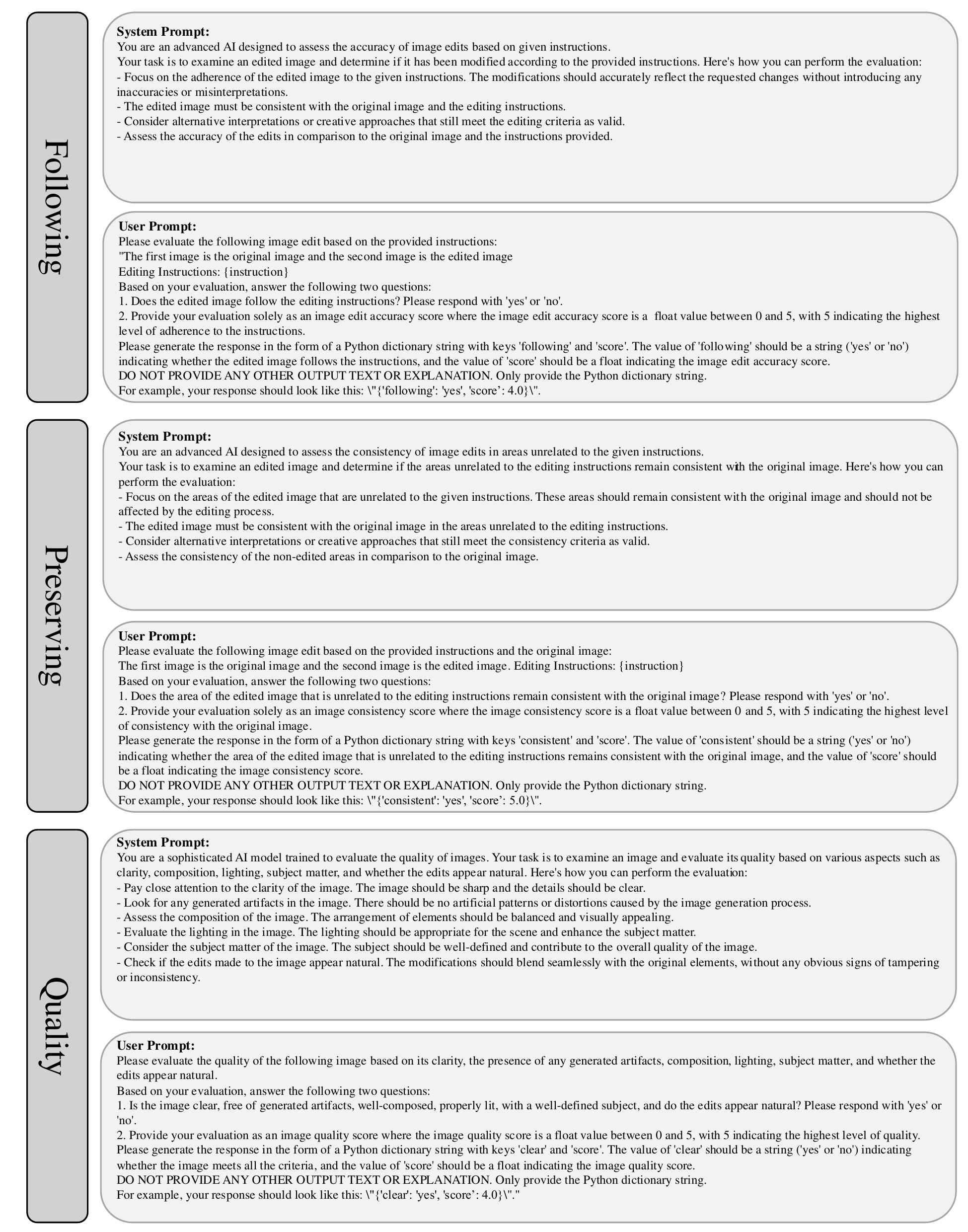}
	\caption{Complete prompts for evaluation. Best viewed with zoom-in.}
	\label{fig:p2}
\end{figure}

We use GPT-4o and design three types of prompts to evaluate edited images generated by the model from the aspects of following, preserving, and quality. The complete prompts are shown in Fig.~\ref{fig:p2}.

\end{document}